\documentclass[letterpaper]{article} 
\usepackage{aaai23}  
\usepackage{times}  
\usepackage{helvet}  
\usepackage{courier}  
\usepackage[hyphens]{url}  
\usepackage{graphicx} 
\usepackage{verbatim}
\usepackage{natbib}  
\usepackage{caption} 
\usepackage{algorithm}
\usepackage{algorithmic}

\usepackage{multirow}
\usepackage{booktabs}
\usepackage{hyperref}

\usepackage{newfloat}
\usepackage{listings}
\DeclareCaptionStyle{ruled}{labelfont=normalfont,labelsep=colon,strut=off} 
\floatstyle{ruled}
\newfloat{listing}{tb}{lst}{}
\floatname{listing}{Listing}
\title{Learning Graph Algorithms With Recurrent Graph Neural Networks}
\author {
    Florian Grötschla\equalcontrib,\textsuperscript{\rm 1}
    Joël Mathys\equalcontrib, \textsuperscript{\rm 1}
    Roger Wattenhofer \textsuperscript{\rm 1}
}
\affiliations {
    \textsuperscript{\rm 1} ETH Zurich\\
    fgroetschla@ethz.ch, jmathys@ethz.ch, wattenhofer@ethz.ch
}

\usepackage{bibentry}

\begin{document}

\maketitle

\begin{abstract}
    
    Classical graph algorithms work well for combinatorial problems that can be thoroughly formalized and abstracted.
    Once the algorithm is derived, it generalizes to instances of any size.
    However, developing an algorithm that handles complex structures and interactions in the real world can be challenging.
    Rather than specifying the algorithm, we can try to learn it from the graph-structured data.
    Graph Neural Networks (GNNs) are inherently capable of working on graph structures; however, they struggle to generalize well, and learning on larger instances is challenging. 
    In order to scale, we focus on a recurrent architecture design that can learn simple graph problems end to end on smaller graphs and then extrapolate to larger instances. 
    As our main contribution, we identify three essential techniques for recurrent GNNs to scale.
    By using (i) skip connections, (ii) state regularization, and (iii) edge convolutions, we can guide GNNs toward extrapolation. 
    This allows us to train on small graphs and apply the same model to much larger graphs during inference.
    Moreover, we empirically validate the extrapolation capabilities of our GNNs on algorithmic datasets.
    
\end{abstract}

\section{Introduction}
We believe that extrapolation is an important milestone for understanding machine learning. Being able to extrapolate to much larger inputs than seen in training is a convincing exhibit for a deeper understanding. For many learning architectures, this is a challenge, because the input size of the architecture is fixed. This is certainly true for multi-layer perceptrons, but even transformers have inputs bounded by a maximum number of tokens.
A prominent exception are Graph Neural Networks (GNNs). In GNNs, each node is operating individually by merely exchanging messages with its neighbors. As such, in principle, GNNs can be trained on small graphs but then run on much larger graphs. Consequently, GNNs seem to be a natural match for studying extrapolation. 

However, GNNs are usually just trained for a fixed and small number of message passing rounds. Nodes can learn a different behavior in each round, which gives them a lot of flexibility to learn even complicated functions. Nevertheless, this flexibility also prevents GNNs from being able to handle problems where information has to be propagated through the whole graph, since these cannot be solved with a fixed number of rounds.

Instead, we want to mimic classical algorithms and be able to adapt the number of rounds a GNN can execute.

To do so, we use a recurrent GNN architecture to learn graph algorithms end-to-end. 
The core idea is to learn only on small graphs during training. Then, for inference, we can adapt the number of convolutions due to the recurrent design and apply more rounds for larger graphs.
While a recurrent architecture seems to be necessary to achieve extrapolation, we show that it is not sufficient. In this paper, we propose measures to guide the model toward extrapolation and get a stable output from the network:
\begin{itemize}
        \item We identify three techniques that lead GNNs towards more stable extrapolation: (i) skip connections to the problem input, (ii) state regularization through L2 loss, and (iii) edge convolutions.
        \item We show that using our approach, it is possible to extra\-polate to larger graph instances, compare our models to existing baselines, and experimentally validate our findings.
\end{itemize}

\section{Related Work}

\textbf{Graph Neural Networks:}
Initially proposed by~\citeauthor{scarselli2008graph}, GNNs have seen a significant popularity boost, and many new architectures have been established~\cite{kipf2016semi, gat_2017}. A key question regards the theoretical expressiveness of such models. 
It has been shown that the expressiveness of traditional GNNs is limited by the WL color refinement algorithm~\cite{xu2018powerful, wl_1968, papp2022theoretical}. 
To achieve maximal expressiveness, message-passing GNNs have to use enough rounds to be able to match the computational power of WL. 
Moreover, increasing the number of rounds was proven to be necessary to solve graph problems~\cite{Loukas2020What}.
In fact, a GNN can provably not solve or, in some cases, even approximate a problem without executing the minimum amount of required rounds~\cite{sato2019approximation}. 
Unfortunately, increasing the number of rounds has been shown to be difficult in practice. The training is more unstable and complex, leading to problems such as oversmoothing and oversquashing \cite{oono2019graph,alon2020bottleneck}. Therefore, most GNNs limit themselves to executing a constant number of rounds~\cite{xu2018powerful}.

\noindent\textbf{Recurrence and Residual Connections:}
To incorporate more layers, residual connections and recurrent neural networks (RNN) have been proposed~\cite{xu2018representation, huang2019residual}. Residual connections are a common technique for building deeper architectures~\cite{he2016deep}. 
Recurrent neural networks are designed to work with variable length input and can keep internal state over long sequences, two examples being LSTMs~\cite{hochreiter1997long} and GRUs~\cite{cho2014properties}. Following this line of work, several GNN architectures have included mechanisms such as gating, residual connections, or reusing the same layer to build deeper models~\cite{li2021training, tang2020towards, li2015gated, huang2019residual,liu2020towards}.

\noindent\textbf{Algorithmic Learning and Extrapolation:} 
The main aim of extrapolation is to solve problems not encountered during training. As such, the differentiable Neural Turing machine~\cite{graves2014neural} or RNNs which generalize to arbitrary input lengths~\cite{gers2001lstm} have been proposed. One notable example is the work by~\citeauthor{NEURIPS2021_3501672e} that presents a recurrent architecture for Convolutional Neural Networks with residual connections. The model can then solve a series of tasks, including mazes, prefix sums, and chess problems.
They show that executing more computation steps enables their networks to achieve excellent extrapolation capabilities on problem instances up to orders of magnitude beyond what was encountered during training.

GNNs have been used to tackle algorithmic problems before. Prominent examples include SAT and TSP or shortest paths through algorithmic alignment~\cite{selsam2018learning, velivckovic2021neural, palm2018recurrent,joshi2022learning}.
Moreover, recent approaches also focus on extrapolation capabilities to larger graphs on algorithmic reasoning problems \cite{tang2020towards,xu2020neural,velivckovic2022clrs,ibarz2022generalist}. However, they only test on graphs slightly larger than those in the training set, resulting in architectures that have difficulty scaling to arbitrary sizes. Contrarily, our work focuses on architectures that can extrapolate to graphs up to 1000 times larger than the ones in the training set.


\section{Model Architecture}
\begin{figure}
  \centering
  \includegraphics[width=\linewidth]{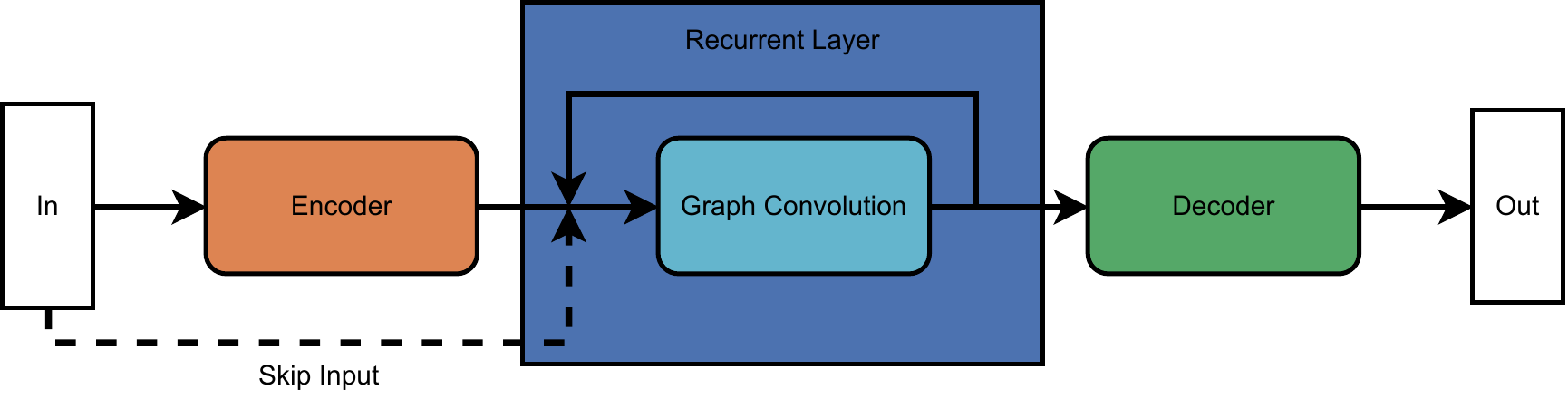}
  \caption{Overview of the recurrent architecture. The recurrent layer consists of a skip connection to the input, and the graph convolution is applied repeatedly. The encoder maps the input to the embedding dimension, while the decoder generates node predictions from the final embedding. The number of executions of the recurrent layer can be varied.}
  \label{fig:model_overview}
\end{figure}

\noindent The recurrent message passing layer is at the core of our architecture, which is illustrated in Figure \ref{fig:model_overview}. It is repeatedly executed on every node of the graph. Therefore, during training, we can use fewer layers and increase the number of layers for larger graphs during inference. The recurrent layer uses a skip connection to the original input, combining the current embedding $h_v^t$ of a node $v$ with the input features and transforming them to a node embedding through a multilayer perceptron (MLP). 
Then, the recurrent graph convolution is applied and uses one round of message passing to derive $h_v^{t+1}$.
The computed embedding is then fed back to the recurrent layer to initiate the next round of computation. 
Furthermore, we use an encoder and decoder MLP to match the input and output dimensions to the recurrent layer's node embedding dimension.

\subsection{Graph Convolutions}
In our work, we use two different graph convolutions: The GIN convolution~\cite{xu2018powerful}, a widely used message passing layer in GNNs, and a GRU convolution~\cite{huang2019residual}, tailored towards the recurrent setting. The two convolutions allow us to compare to what extent a recurrent convolution alone is sufficient or if a more specialized convolution, such as the GRU, is required.
Note that at every timestep, the GRU takes two inputs, the output of the convolution and its previous state. We use variations of both convolutions by adding an MLP on the edges before the aggregation. We refer to the versions without MLP as RecGIN and RecGRU and the ones with edge convolution as RecGIN-E and RecGRU-E.
We define the RecGIN-E update as:
\begin{displaymath}
    h_v^{t+1} = \Theta_1\left((1 + \epsilon) \cdot h_v^t + \sum_{w \in N(v)} \Theta_2(h_v^t \| h_w^t)\right)
\end{displaymath}
where $\|$ denotes concatenation and RecGRU-E is defined as:
\begin{displaymath}
    h_v^{t+1} = \textsc{GRU}\left(\left(\sum_{w \in N(v)} \Theta(h_v^t \| h_w^t)\right), h_v^t\right)
\end{displaymath}
The only difference to RecGIN and RecGRU is the additional MLP for the edges.

\subsection{Extrapolation Techniques}
The recurrent architecture allows us to vary the number of computation steps. This is not yet sufficient to achieve extrapolation. The predictions should also stabilize, so the GNN can use the information propagated through the additional layers without digressing.
To guide the model toward extrapolation, we identify three essential techniques.

First, we introduce skip connections from the original input to the beginning of each recurrent layer as \citeauthor{NEURIPS2021_3501672e} proposed.
They observed that when the number of recurrent layers increases, the network tends to ``forget'' its initial features.
By explicitly passing the input features to every execution of the recurrent layer, the network can recall its initial state and the problem it has to solve.

Another problem concerns the stability of computed embeddings. Once a solution is found by the GNN, executing more rounds should not change the prediction anymore.
However, even slight deviations of the embedding in one layer can magnify in the following layers.
We found that adding regularization in the form of an L2 loss on the embeddings improves the stability of the computation. 

Finally, we add an MLP on pairs of node embeddings for every edge. 
The MLP allows the GNN to differentiate between neighbors and choose what messages to include in the aggregation step over all neighbors. This enables better control over information propagation through the graph and improves extrapolation capabilities.

\subsection{Tasks}
We test our model on multiple synthetic datasets that are specifically tailored to evaluate the ability of a GNN to gather and combine information over long distances. 

\textit{Path Finding:}
Given a tree, predict if a node lies on the path between two marked nodes. 

\textit{Prefix Sum:}
Paths are given where every node either has a one or zero as its initial feature. For each node, the sum of all initial features to its left modulo two has to be predicted.

\textit{Distance:}
Given a sparse graph with a marked starting node, for each node, predict the distance of the shortest path to the starting node modulo 2. 


\section{Experimental Evaluation}

\begin{figure}
  \centering
  \includegraphics[width=\linewidth]{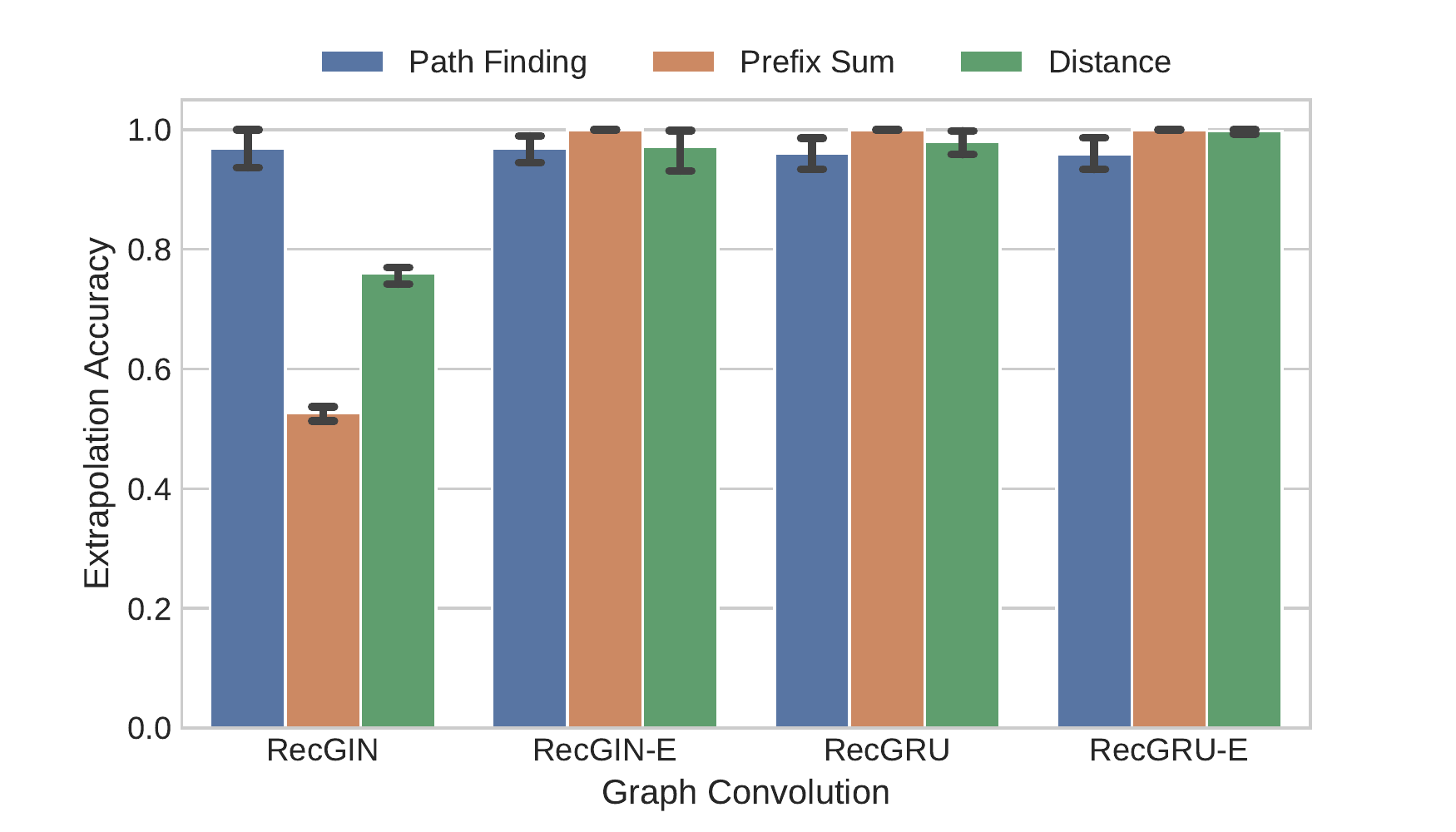}
  \caption{Comparison of extrapolation accuracy for different graph convolutions. Adapting the convolution on the edges or using a GRU are beneficial for extrapolation.}
  \label{fig:convolution-comparison}
\end{figure}
\noindent 
All models were trained for 100 epochs on graphs of size 10 using 12 instances of the recurrent layer. 
To evaluate extrapolation, we use graphs of size 100 and execute 120 rounds. 
For comparisons, we consider the model with the best loss on a validation dataset with graphs of size 10 that were not used for training.
All values are averaged over 5 runs.
The code to reproduce the experiments has been made available online~\footnote{\url{https://github.com/floriangroetschla/Recurrent-GNNs-for-algorithm-learning}}.

\subsection{Model Architecture}
First, we compare the different graph convolutions and the effect of including an edge convolution. 
We train models with skip connections and regularization on graphs of size 10 and evaluate their accuracy on graphs of size 100.
As illustrated in Figure~\ref{fig:convolution-comparison}, we observe that models using a GRU in the convolution perform better than the GIN variants.
Adding the edge convolution results in better generalization accuracy overall, although only marginally for RecGRU.
Models that use the edge convolution can reach perfect accuracy, while this is not the case for RecGIN.
Therefore, we conclude that the MLP on edges is helpful, albeit not completely necessary for all convolutions.

\begin{figure}
  \centering
  \includegraphics[width=\linewidth]{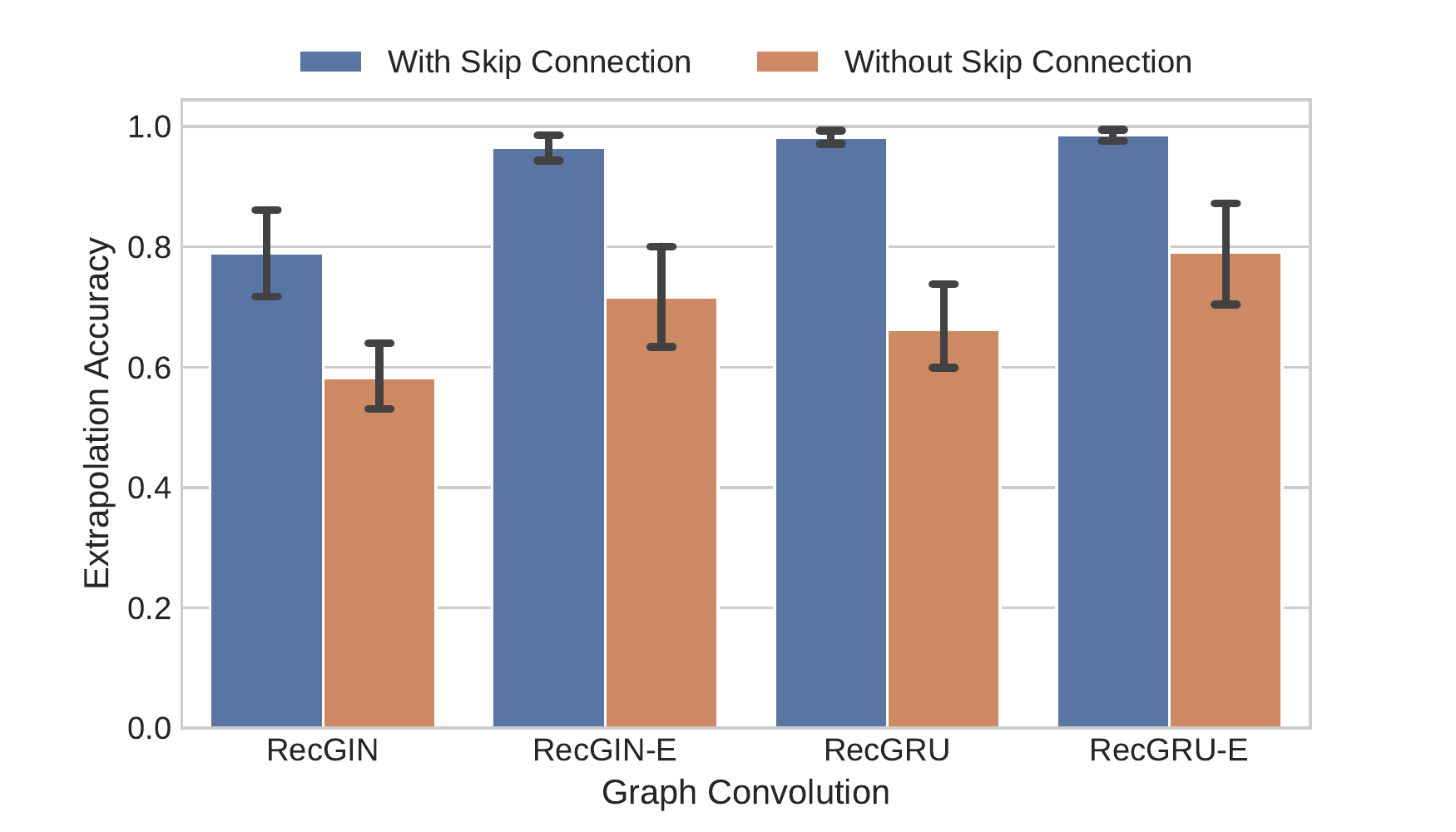}
  \caption{Extrapolation accuracy for each graph convolution for the Prefix Sum task. We hypothesize that the ability to recall the initial features is crucial for extrapolation.} 
  \label{fig:skip-input}
\end{figure}

Regarding the use of skip connections, we can observe a meaningful difference, shown exemplarily on the prefix task in Figure~\ref{fig:skip-input}.
Even models that use edge convolutions are not able to reach good accuracy without them.
This confirms the findings of~\citeauthor{NEURIPS2021_3501672e}, that access to the input features is a crucial part of gaining extrapolation abilities. We conclude that adding the skip connection to the inputs is also an excellent tool for extrapolation for GNNs. 




\subsection{Stabilization and Extrapolation}
\begin{figure}
  \centering
  \includegraphics[width=\linewidth]{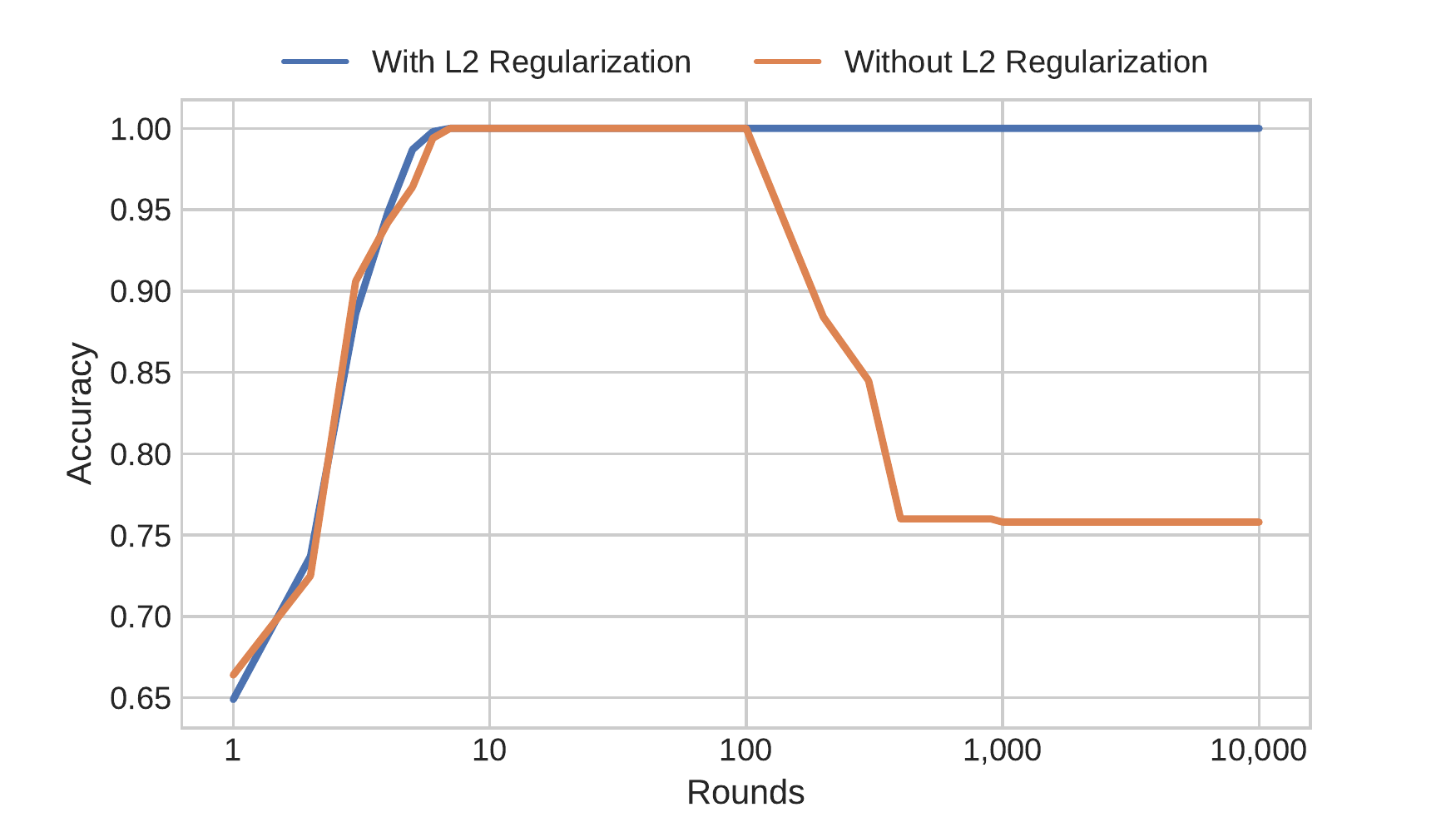}
  \caption{Accuracy for the Distance task on graphs of size 10 for RecGRU-E over the number of rounds the GNN executes. Initially, the accuracy increases as information flows through the graph. Then, predictions stabilize for models using regularization while it decreases for models without it. 
  }
  \label{fig:stabilization}
\end{figure}
Before, we tested extrapolation to graphs of size 100. We want to extend this to even larger graphs.
In order for a model to extrapolate, it also needs to be able to stabilize predictions when many convolutions are applied.
We evaluate the stabilization on graphs of size 10 and then increase the number of layers from 10 up to 10,000.
Figure~\ref{fig:stabilization} shows the effect of using L2 regularization for the distance task.
Within the first 10 rounds, the accuracy increases as the information can propagate further through the graph.
Then, both versions output correct predictions until approximately 100 layers. Afterwards, the models trained with L2 regularization still stay stable while accuracy declines rapidly without regularization.
We hypothesize that slight deviations in the node embeddings can lead to the performance degradation over time. 
The additional regularization loss incentivizes the model to keep embeddings compact. We conclude that without regularization, the model can still exhibit extrapolation to some degree. However, for further extrapolation, it is essential to use regularization.

\begin{table}[]
  \resizebox{\columnwidth}{!}{%

\begin{tabular}{clc||ccccc}
    &Model & $n = 10$ & $n = 50$ & $n = 100$ & $n = 1,000$ & $n = 10,000$ \\
    \midrule
\parbox[t]{4mm}{\multirow{4}{*}{\rotatebox[origin=c]{90}{\parbox{1cm}{\small \centering Path \ Finding}}}}

&GIN  & $\mathbf{1.00 \pm 0.00}$ & $0.70 \pm 0.10$ & $0.52 \pm 0.11$ & $0.22 \pm 0.07$ & $0.08 \pm 0.04$\\
&IterGNN  & $0.86 \pm 0.09$ & $0.51 \pm 0.12$ & $0.41 \pm 0.08$ & $0.15 \pm 0.06$ & $0.06 \pm 0.02$\\
&RecGIN-E  & $\mathbf{1.00 \pm 0.00}$ & $0.91 \pm 0.04$ & $0.88 \pm 0.09$ & $0.66 \pm 0.26$ & $0.33 \pm 0.31$\\
&RecGRU-E  & $\mathbf{1.00 \pm 0.00}$ & $\mathbf{0.96 \pm 0.09}$ & $\mathbf{0.92 \pm 0.16}$ & $\mathbf{0.86 \pm 0.30}$ & $\mathbf{0.81 \pm 0.39}$\\
\midrule
\parbox[t]{4mm}{\multirow{4}{*}{\rotatebox[origin=c]{90}{\parbox{1cm}{\small \centering Prefix \ Sum}}}}

&GIN  & $0.95 \pm 0.12$ & $0.54 \pm 0.11$ & $0.48 \pm 0.16$ & $0.41 \pm 0.22$ & $0.40 \pm 0.22$\\
&IterGNN  & $\mathbf{1.00 \pm 0.00}$ & $0.58 \pm 0.11$ & $0.51 \pm 0.17$ & $0.40 \pm 0.25$ & $0.40 \pm 0.26$\\
&RecGIN-E  & $\mathbf{1.00 \pm 0.00}$ & $\mathbf{1.00 \pm 0.00}$ & $\mathbf{1.00 \pm 0.00}$ & $\mathbf{1.00 \pm 0.00}$ & $\mathbf{1.00 \pm 0.00}$\\
&RecGRU-E  & $\mathbf{1.00 \pm 0.00}$ & $\mathbf{1.00 \pm 0.00}$ & $\mathbf{1.00 \pm 0.00}$ & $\mathbf{1.00 \pm 0.00}$ & $0.98 \pm 0.03$\\
\midrule
\parbox[t]{1mm}{\multirow{4}{*}{\rotatebox[origin=c]{90}{\parbox{1cm}{\small \centering Distance}}}}

&GIN  & $\mathbf{1.00 \pm 0.00}$ & $0.89 \pm 0.05$ & $0.83 \pm 0.08$ & $0.59 \pm 0.20$ & $0.52 \pm 0.23$\\
&IterGNN  & $\mathbf{1.00 \pm 0.00}$ & $0.99 \pm 0.01$ & $0.98 \pm 0.04$ & $0.81 \pm 0.05$ & $0.57 \pm 0.17$\\
&RecGIN-E  & $\mathbf{1.00 \pm 0.00}$ & $0.98 \pm 0.02$ & $0.91 \pm 0.04$ & $0.74 \pm 0.07$ & $0.57 \pm 0.08$\\
&RecGRU-E  & $\mathbf{1.00 \pm 0.00}$ & $\mathbf{1.00 \pm 0.00}$ & $\mathbf{1.00 \pm 0.00}$ & $\mathbf{0.99 \pm 0.02}$ & $\mathbf{0.97 \pm 0.04}$\\

\end{tabular}

  }
  \caption{We report the F1 score for the evaluation of extrapolation abilities. All models were only trained on graphs of size 10. The non-recurrent GIN can not extrapolate. The RecGRU-E outperforms IterGNN and can extrapolate well to graphs that are 1,000 times larger than the graphs encountered during training. 
  }
  \label{tab:extrapolation}

\end{table}

Lastly, we evaluate our model's extrapolation capabilities by testing it on even larger graphs and comparing it against existing baselines as illustrated in Table~\ref{tab:extrapolation}. All listed models were trained on graphs of size $10$. We compare against a non-recurrent GIN baseline model that uses 10 separate GIN convolutions with skip connections and was trained with L2 regularization. As the architecture is not recurrent, it always uses exactly 10 rounds of message passing. We observe that GIN can still solve the task on graph sizes it has encountered during training. However, even for slightly larger graphs, it struggles to extrapolate as it can not propagate the relevant information far enough to solve the task. Furthermore, we compare our models to IterGNN~\cite{tang2020towards}. IterGNN is also based on a recurrent architecture and can vary the number of iterations. In addition, it learns a stopping criterion. Even though it can extrapolate to some degree, it can not extend to very large graphs. Both the RecGIN-E and RecGRU-E can extrapolate to graphs that are up to 1000 times larger than the graphs encountered during training, with the GRU variant achieving the best results overall.


\subsection{Learned Algorithms}
\begin{figure}
  \centering
  \includegraphics[trim={3cm 2cm 2cm 2cm}, clip, width=\linewidth]{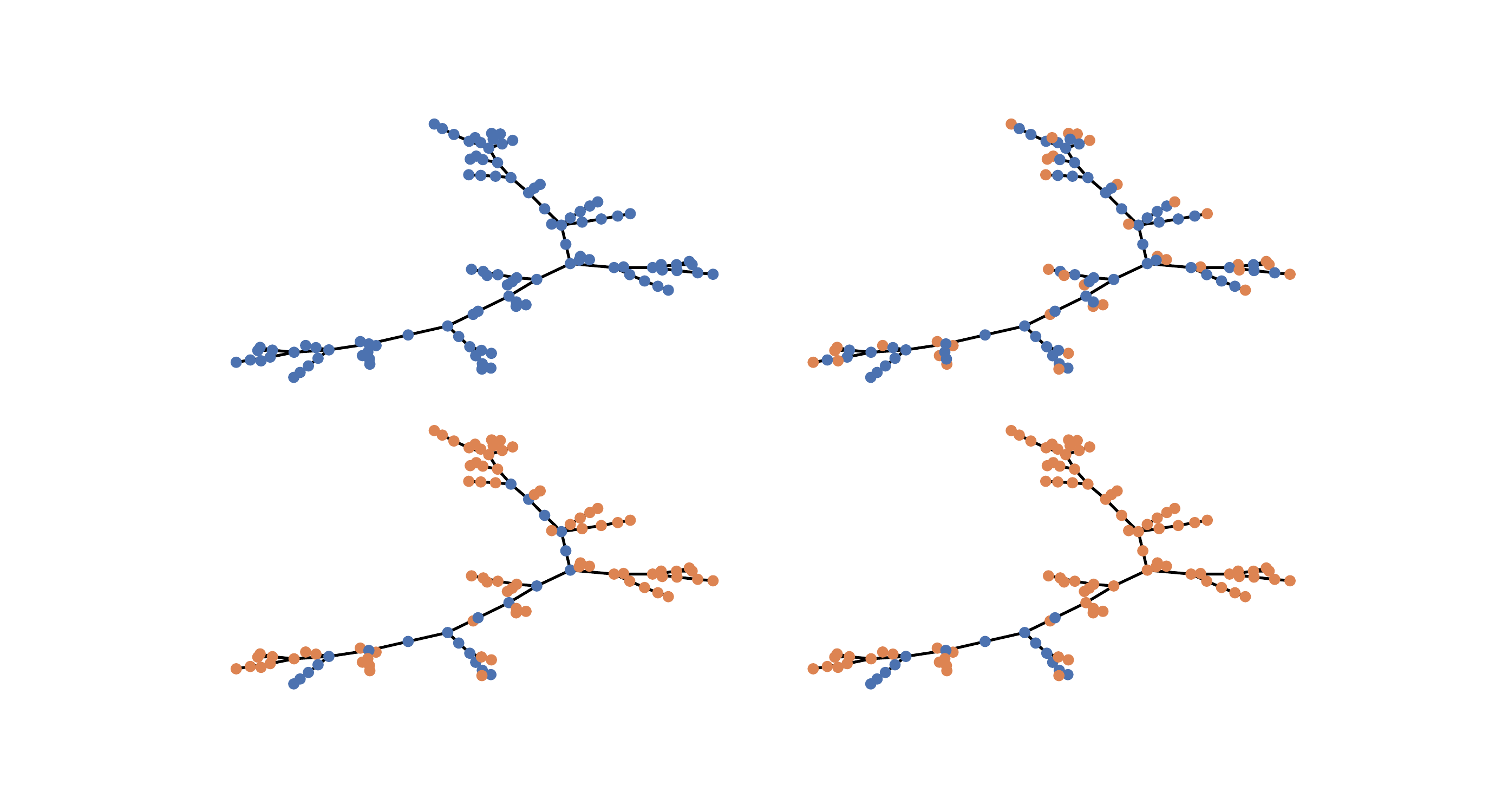}
   \caption{Development of the predictions for the Path Finding task after executing more rounds, from top left to bottom right (row by row). The GNN  scales to different graph sizes by successively removing nodes that are not part of the path.}

  \label{fig:visualization}
\end{figure}
To further evaluate to what extent our model could learn algorithmic behavior, we plot predictions after several iterations of the recurrent layer for the Path Finding task. 
By visually inspecting the outputs, we can observe that the GNN uses a dead-end-filling algorithm that ``removes'' nodes that can not be part of the path until it reaches the path between the endpoints.
The algorithm is correct and scales to arbitrary graph sizes.
Similarly, on the Distance task, the GNN emulates a Breadth First Search while information for the total sum is propagated from left to right for the Prefix Sum task. 

\section{Conclusion}
Classic graph algorithms can scale to any graph size by allowing the algorithm to execute more computation steps for larger instances. We want to mimic this ability by training on small graphs and then extrapolate to larger instances. 
Traditional GNN architectures can only execute a fixed number of rounds. To make up for this shortcoming, we use a recurrent architecture that can vary this number.

In addition to recurrence, we take additional measures to make our models scale.
Skip connections to the input features, the addition of MLPs on edges and the regularization of node embeddings help GNNs to extrapolate well. 
With our approach, we can train on small graph instances and extrapolate to graphs outside the training regime that are orders of magnitudes larger. 
We evaluate our models on algorithmic datasets and compare them to existing baselines. Using the techniques above, our model can extrapolate to graphs 1,000 times larger than during training. Moreover, the computed solutions stabilize and do not change after executing more rounds. 

While we indicate essential tools to help extrapolation, more work is required toward algorithmic learning on graphs.
Possible directions include discrete embeddings to imitate state machine or symbolic algorithms that go beyond just defining state transitions.
This could lead to interpretable algorithmic behavior and close the gap between human and artificial decision-making.

\pagebreak

\bibliography{refs}

\begin{thebibliography}{29}
\providecommand{\natexlab}[1]{#1}

\bibitem[{Alon and Yahav(2020)}]{alon2020bottleneck}
Alon, U.; and Yahav, E. 2020.
\newblock On the bottleneck of graph neural networks and its practical
  implications.
\newblock \emph{arXiv preprint arXiv:2006.05205}.

\bibitem[{Cho et~al.(2014)Cho, Van~Merri{\"e}nboer, Bahdanau, and
  Bengio}]{cho2014properties}
Cho, K.; Van~Merri{\"e}nboer, B.; Bahdanau, D.; and Bengio, Y. 2014.
\newblock On the properties of neural machine translation: Encoder-decoder
  approaches.
\newblock \emph{arXiv preprint arXiv:1409.1259}.

\bibitem[{Gers and Schmidhuber(2001)}]{gers2001lstm}
Gers, F.~A.; and Schmidhuber, J. 2001.
\newblock LSTM recurrent networks learn simple context-free and
  context-sensitive languages.
\newblock \emph{IEEE Transactions on Neural Networks}, 12(6): 1333--1340.

\bibitem[{Graves, Wayne, and Danihelka(2014)}]{graves2014neural}
Graves, A.; Wayne, G.; and Danihelka, I. 2014.
\newblock Neural turing machines.
\newblock \emph{arXiv preprint arXiv:1410.5401}.

\bibitem[{He et~al.(2016)He, Zhang, Ren, and Sun}]{he2016deep}
He, K.; Zhang, X.; Ren, S.; and Sun, J. 2016.
\newblock Deep residual learning for image recognition.
\newblock In \emph{Proceedings of the IEEE conference on computer vision and
  pattern recognition}, 770--778.

\bibitem[{Hochreiter and Schmidhuber(1997)}]{hochreiter1997long}
Hochreiter, S.; and Schmidhuber, J. 1997.
\newblock Long short-term memory.
\newblock \emph{Neural computation}, 9(8): 1735--1780.

\bibitem[{Huang and Carley(2019)}]{huang2019residual}
Huang, B.; and Carley, K.~M. 2019.
\newblock Residual or gate? towards deeper graph neural networks for inductive
  graph representation learning.
\newblock \emph{arXiv preprint arXiv:1904.08035}.

\bibitem[{Ibarz et~al.(2022)Ibarz, Kurin, Papamakarios, Nikiforou, Bennani,
  Csord{\'a}s, Dudzik, Bo{\v{s}}njak, Vitvitskyi, Rubanova
  et~al.}]{ibarz2022generalist}
Ibarz, B.; Kurin, V.; Papamakarios, G.; Nikiforou, K.; Bennani, M.;
  Csord{\'a}s, R.; Dudzik, A.; Bo{\v{s}}njak, M.; Vitvitskyi, A.; Rubanova, Y.;
  et~al. 2022.
\newblock A Generalist Neural Algorithmic Learner.
\newblock \emph{arXiv preprint arXiv:2209.11142}.

\bibitem[{Joshi et~al.(2022)Joshi, Cappart, Rousseau, and
  Laurent}]{joshi2022learning}
Joshi, C.~K.; Cappart, Q.; Rousseau, L.-M.; and Laurent, T. 2022.
\newblock Learning the travelling salesperson problem requires rethinking
  generalization.
\newblock \emph{Constraints}, 1--29.

\bibitem[{Kipf and Welling(2016)}]{kipf2016semi}
Kipf, T.~N.; and Welling, M. 2016.
\newblock Semi-supervised classification with graph convolutional networks.
\newblock \emph{arXiv preprint arXiv:1609.02907}.

\bibitem[{Leman and Weisfeiler(1968)}]{wl_1968}
Leman, A.; and Weisfeiler, B. 1968.
\newblock A reduction of a graph to a canonical form and an algebra arising
  during this reduction.
\newblock \emph{Nauchno-Technicheskaya Informatsiya}, 2(9): 12--16.

\bibitem[{Li et~al.(2021)Li, M{\"u}ller, Ghanem, and Koltun}]{li2021training}
Li, G.; M{\"u}ller, M.; Ghanem, B.; and Koltun, V. 2021.
\newblock Training graph neural networks with 1000 layers.
\newblock In \emph{International conference on machine learning}, 6437--6449.
  PMLR.

\bibitem[{Li et~al.(2015)Li, Tarlow, Brockschmidt, and Zemel}]{li2015gated}
Li, Y.; Tarlow, D.; Brockschmidt, M.; and Zemel, R. 2015.
\newblock Gated graph sequence neural networks.
\newblock \emph{arXiv preprint arXiv:1511.05493}.

\bibitem[{Liu, Gao, and Ji(2020)}]{liu2020towards}
Liu, M.; Gao, H.; and Ji, S. 2020.
\newblock Towards deeper graph neural networks.
\newblock In \emph{Proceedings of the 26th ACM SIGKDD international conference
  on knowledge discovery \& data mining}, 338--348.

\bibitem[{Loukas(2020)}]{Loukas2020What}
Loukas, A. 2020.
\newblock What graph neural networks cannot learn: depth vs width.
\newblock In \emph{International Conference on Learning Representations}.

\bibitem[{Oono and Suzuki(2019)}]{oono2019graph}
Oono, K.; and Suzuki, T. 2019.
\newblock Graph neural networks exponentially lose expressive power for node
  classification.
\newblock \emph{arXiv preprint arXiv:1905.10947}.

\bibitem[{Palm, Paquet, and Winther(2018)}]{palm2018recurrent}
Palm, R.; Paquet, U.; and Winther, O. 2018.
\newblock Recurrent relational networks.
\newblock \emph{Advances in Neural Information Processing Systems}, 31.

\bibitem[{Papp and Wattenhofer(2022)}]{papp2022theoretical}
Papp, P.~A.; and Wattenhofer, R. 2022.
\newblock A Theoretical Comparison of Graph Neural Network Extensions.
\newblock \emph{arXiv preprint arXiv:2201.12884}.

\bibitem[{Sato, Yamada, and Kashima(2019)}]{sato2019approximation}
Sato, R.; Yamada, M.; and Kashima, H. 2019.
\newblock Approximation ratios of graph neural networks for combinatorial
  problems.
\newblock \emph{Advances in Neural Information Processing Systems}, 32.

\bibitem[{Scarselli et~al.(2008)Scarselli, Gori, Tsoi, Hagenbuchner, and
  Monfardini}]{scarselli2008graph}
Scarselli, F.; Gori, M.; Tsoi, A.~C.; Hagenbuchner, M.; and Monfardini, G.
  2008.
\newblock The graph neural network model.
\newblock \emph{IEEE transactions on neural networks}, 20(1): 61--80.

\bibitem[{Schwarzschild et~al.(2021)Schwarzschild, Borgnia, Gupta, Huang,
  Vishkin, Goldblum, and Goldstein}]{NEURIPS2021_3501672e}
Schwarzschild, A.; Borgnia, E.; Gupta, A.; Huang, F.; Vishkin, U.; Goldblum,
  M.; and Goldstein, T. 2021.
\newblock Can You Learn an Algorithm? Generalizing from Easy to Hard Problems
  with Recurrent Networks.
\newblock In \emph{Advances in Neural Information Processing Systems},
  volume~34, 6695--6706. Curran Associates, Inc.

\bibitem[{Selsam et~al.(2018)Selsam, Lamm, B{\"u}nz, Liang, de~Moura, and
  Dill}]{selsam2018learning}
Selsam, D.; Lamm, M.; B{\"u}nz, B.; Liang, P.; de~Moura, L.; and Dill, D.~L.
  2018.
\newblock Learning a SAT solver from single-bit supervision.
\newblock \emph{arXiv preprint arXiv:1802.03685}.

\bibitem[{Tang et~al.(2020)Tang, Huang, Gu, Lu, and Su}]{tang2020towards}
Tang, H.; Huang, Z.; Gu, J.; Lu, B.-L.; and Su, H. 2020.
\newblock Towards scale-invariant graph-related problem solving by iterative
  homogeneous gnns.
\newblock \emph{Advances in Neural Information Processing Systems}, 33:
  15811--15822.

\bibitem[{Veli{\v{c}}kovi{\'c} et~al.(2022)Veli{\v{c}}kovi{\'c}, Badia, Budden,
  Pascanu, Banino, Dashevskiy, Hadsell, and Blundell}]{velivckovic2022clrs}
Veli{\v{c}}kovi{\'c}, P.; Badia, A.~P.; Budden, D.; Pascanu, R.; Banino, A.;
  Dashevskiy, M.; Hadsell, R.; and Blundell, C. 2022.
\newblock The CLRS Algorithmic Reasoning Benchmark.
\newblock \emph{arXiv preprint arXiv:2205.15659}.

\bibitem[{Veli{\v{c}}kovi{\'c} and Blundell(2021)}]{velivckovic2021neural}
Veli{\v{c}}kovi{\'c}, P.; and Blundell, C. 2021.
\newblock Neural algorithmic reasoning.
\newblock \emph{Patterns}, 2(7): 100273.

\bibitem[{Veli{\v{c}}kovi{\'c} et~al.(2017)Veli{\v{c}}kovi{\'c}, Cucurull,
  Casanova, Romero, Lio, and Bengio}]{gat_2017}
Veli{\v{c}}kovi{\'c}, P.; Cucurull, G.; Casanova, A.; Romero, A.; Lio, P.; and
  Bengio, Y. 2017.
\newblock Graph attention networks.
\newblock \emph{arXiv preprint arXiv:1710.10903}.

\bibitem[{Xu et~al.(2018{\natexlab{a}})Xu, Hu, Leskovec, and
  Jegelka}]{xu2018powerful}
Xu, K.; Hu, W.; Leskovec, J.; and Jegelka, S. 2018{\natexlab{a}}.
\newblock How powerful are graph neural networks?
\newblock \emph{arXiv preprint arXiv:1810.00826}.

\bibitem[{Xu et~al.(2018{\natexlab{b}})Xu, Li, Tian, Sonobe, Kawarabayashi, and
  Jegelka}]{xu2018representation}
Xu, K.; Li, C.; Tian, Y.; Sonobe, T.; Kawarabayashi, K.-i.; and Jegelka, S.
  2018{\natexlab{b}}.
\newblock Representation learning on graphs with jumping knowledge networks.
\newblock In \emph{International Conference on Machine Learning}, 5453--5462.
  PMLR.

\bibitem[{Xu et~al.(2020)Xu, Zhang, Li, Du, Kawarabayashi, and
  Jegelka}]{xu2020neural}
Xu, K.; Zhang, M.; Li, J.; Du, S.~S.; Kawarabayashi, K.-i.; and Jegelka, S.
  2020.
\newblock How neural networks extrapolate: From feedforward to graph neural
  networks.
\newblock \emph{arXiv preprint arXiv:2009.11848}.

\end{thebibliography}

\end{document}